\documentclass[sigconf]{acmart}

\acmConference[ICSE 2026]{48th International Conference on Software Engineering}{April 2026}{Rio de Janeiro, Brazil}
\AtBeginDocument{%
  }

\setcopyright{acmlicensed}
\copyrightyear{2018}
\acmYear{2018}
\acmDOI{XXXXXXX.XXXXXXX}
\usepackage{xspace}
\usepackage{enumitem}
\acmISBN{978-1-4503-XXXX-X/2018/06}

\begin{document}

\title{90\% Faster, 100\% Code-Free: MLLM-Driven Zero-Code 3D Game Development}

\author{Runxin Yang}
\affiliation{%
  \institution{The Chinese University of Hong Kong}
  \city{Hong Kong}
  \country{China}}
\email{1155211029@link.cuhk.edu.hk}

\author{Yuxuan Wan}
\authornote{Project lead.}
\affiliation{%
  \institution{The Chinese University of Hong Kong}
  \city{Hong Kong}
  \country{China}}
\email{yxwan@link.cuhk.edu.hk}

\author{Shuqing Li}
\authornote{Shuqing Li is the corresponding author.}
\affiliation{%
  \institution{The Chinese University of Hong Kong}
  \city{Hong Kong}
  \country{China}}
\email{sqli21@cse.cuhk.edu.hk}

\author{Michael R. Lyu}
\affiliation{%
  \institution{The Chinese University of Hong Kong}
  \city{Hong Kong}
  \country{China}}
\email{lyu@cse.cuhk.edu.hk}

\renewcommand{\shortauthors}{Yang et al.}
\newcommand{\yx}[1]{\textcolor{blue}{#1}}
\newcommand{\sq}[1]{\textcolor{magenta}{#1}}
\newcommand{\sqcomment}[1]{\textcolor{magenta}{[Shuqing: #1]}}
\newcommand{\methodname}{UniGen\xspace}

\begin{abstract}
Developing 3D games requires specialized expertise across multiple domains, including programming, 3D modeling, and engine configuration, which limits access to millions of potential creators.
Recently, researchers have begun to explore automated game development.
However, existing approaches face three primary challenges: (1) limited scope to 2D content generation or isolated code snippets; (2) requirement for manual integration of generated components into game engines; and (3) poor performance on handling interactive game logic and state management.
While Multimodal Large Language Models (MLLMs) demonstrate potential capabilities to ease the game generation task, a critical gap still remains in translating these outputs into production-ready, executable game projects based on game engines such as Unity and Unreal Engine. 

To bridge the gap, this paper introduces \methodname, the first end-to-end coordinated multi-agent framework that automates zero-coding development of runnable 3D games from natural language requirements. 
Specifically, \methodname uses a Planning Agent that interprets user requirements into structured blueprints and engineered logic descriptions; after which a Generation Agent produces executable C\# scripts; then an Automation Agent handles engine-specific component binding and scene construction; and lastly a Debugging Agent provides real-time error correction through conversational interaction.
We evaluated \methodname on three distinct game prototypes. Results demonstrate that \methodname not only democratizes game creation by requiring no coding from the user, but also reduces development time by 91.4\%.  We release \methodname at 
\url{https://github.com/yxwan123/UniGen}. A video demonstration is available at \url{https://www.youtube.com/watch?v=xyJjFfnxUx0}.
\end{abstract}



\maketitle

\section{Introduction}
The development of 3D games demands expertise across multiple disciplines, including programming, computer graphics, physics simulation, and interactive design. This creates substantial barriers that exclude millions of potential creators from the \$200+ billion gaming industry~\cite{fortune2025gameengine}. 
While recent advances in AI-assisted development~\cite{Google_AIMeetsGames} have shown promise in automating specific aspects of game creation, current approaches remain fundamentally limited.

Existing AI-assisted game development systems face critical limitations. Prior work has primarily focused on 2D game generation (e.g., procedural level design) or partial automation (e.g., NPC behavior scripting), while comprehensive 3D game development, which demands sophisticated spatial reasoning, physics interactions, and engine compatibility, remains underexplored. Current systems: (1) generate only isolated components such as code fragments or assets that fail to constitute complete game systems~\cite{guzdial2022pcgml}; (2) require extensive manual integration into production game engines, with studies showing over 70\% of development time consumed by configuration and component binding tasks~\cite{UnityGamingReport2024}; and (3) exhibit limited capability in synthesizing complex interactive logic, achieving less than 50\% accuracy on gameplay benchmarks~\cite{zhang2023gameeval}. For instance, SceneDreamer~\cite{Chen2023SceneDreamerU3} can synthesize 3D landscapes but lacks integration with gameplay mechanics, while GaML~\cite{Herzig2013GAML} still requires manual coding for implementation. These inefficiencies represent a significant opportunity for automation to transform game development practices.

The rapid advancement of Multimodal Large Language Models (MLLMs) has demonstrated their strong capabilities for automating complex creative task,
including code generation~\cite{li2024mmcode, zhao2025chartcoder}. 
This creates an unprecedented opportunity for zero-code 3D game development. 
In this work, we present \methodname, an end-to-end system that leverages MLLMs to generate complete, playable 3D games directly from natural language descriptions. 
Our vision is to democratize game creation by enabling non-programmers (e.g., designers, educators, and hobbyists) to develop functional 3D games without writing code or manually configuring game engines.
With \methodname, users simply describe their desired game in natural language (e.g., ``Create a 3D platformer with moving platforms and enemies''), and the system automatically generates a complete, playable Unity project. 
We focus on Unity as our implementation platform due to its dominant market position, powering over 60\% of mobile games~\cite{UnityGamingReport2024} and half of all PC and console games~\cite{UnityGamingReport2024}. Although our implementation specifically targets Unity, the underlying multi-agent framework and pipeline design are engine-agnostic and can be generalized to other game development platforms.

To realize this vision, \methodname addresses three core technical challenges: (1) \emph{MLLM-engine integration}, bridging the gap between high-level natural language prompts and executable Unity workflows through editor scripting APIs; (2) \emph{spatial and logical reasoning}, enhancing MLLMs to correctly handle 3D mechanics such as collision detection and physics-based interactions; and (3) \emph{end-to-end automation}, constructing a pipeline that transforms natural language descriptions into playable builds without manual intervention. Our experiments demonstrate that \methodname achieves over 90\% functional completeness in generating diverse 3D game prototypes while reducing development time by 91.4\%, from hours to under ten minutes. These results underscore the transformative potential of \methodname in reshaping software engineering practices for game development, offering both efficiency gains and new pathways for democratizing access to 3D software creation.

\section{Background and Related Work}
The automation of game development has been a longstanding goal in software engineering, motivated by the complexity and multidisciplinary nature of modern game creation. This section reviews existing approaches across three main dimensions: code generation, visual content synthesis, and development tool support.

Early efforts in AI-assisted game development have predominantly addressed 2D procedural content generation. Systems such as ANGELINA~\cite{cook2016angelina} and MarioGPT~\cite{sudhakaran2023mariogpt} demonstrated the ability to create levels and mechanics through rule-based or evolutionary algorithms automatically. While effective for simple spatial structures, these approaches remain limited in scope and cannot generalize to the complexity of 3D environments. More recently, MLLMs such as GPT-4.1~\cite{openai2025gpt41}, Gemini-2.5~\cite{kavukcuoglu2025gemini25} have demonstrated capabilities in automating specific code generation tasks by generating functional Unity C\# scripts. However, these tools primarily address script generation in isolation, where generated scripts must be manually integrated into Unity projects.
This integration requires resolving engine-specific dependencies and configuring project structures via the Unity Editor, demanding substantial technical expertise. Our empirical evaluation reveals that over half of the “AI-assisted” development time is still spent on manual integration tasks, resulting in limited productivity improvements.

Another line of work is Interactive Video Generation~\cite{wu2024language, blattmann2023stable}, which uses techniques to generate dynamic visual content based on textual instructions. However, these methods primarily focus on visual presentation rather than interactive, rule-based game logic and state management. They are unable to generate playable game experiences that include player input responses or goal-driven mechanisms. Therefore, while they may have potential as an aid for asset creation, they don't solve the core challenge of automating the construction of functional gameplay.

Beyond visual generation, specialized tools for game development, such as Unity's ML-Agents~\cite{juliani2018unity} platform, provide an environment for training agent behaviors. However, they still heavily rely on manual coding to define the training environment and game rules. Similarly, general AI explainability tools like Google’s What-If Tool~\cite{wexler2019what} allow for hypothetical analysis of model behavior but are not designed for generating or debugging complete game logic within a game engine. These tools are insufficient in achieving an end-to-end automation from concept to finished product.

In contrast to these existing approaches, our work presents the first comprehensive solution that addresses the end-to-end automation challenge. Unlike systems that generate isolated components, \methodname integrates four specialized agents to 
bridge the semantic gap between natural language specifications and executable Unity projects.

\section{Methodology}
\begin{figure}
    \centering
    \includegraphics[width=\columnwidth]{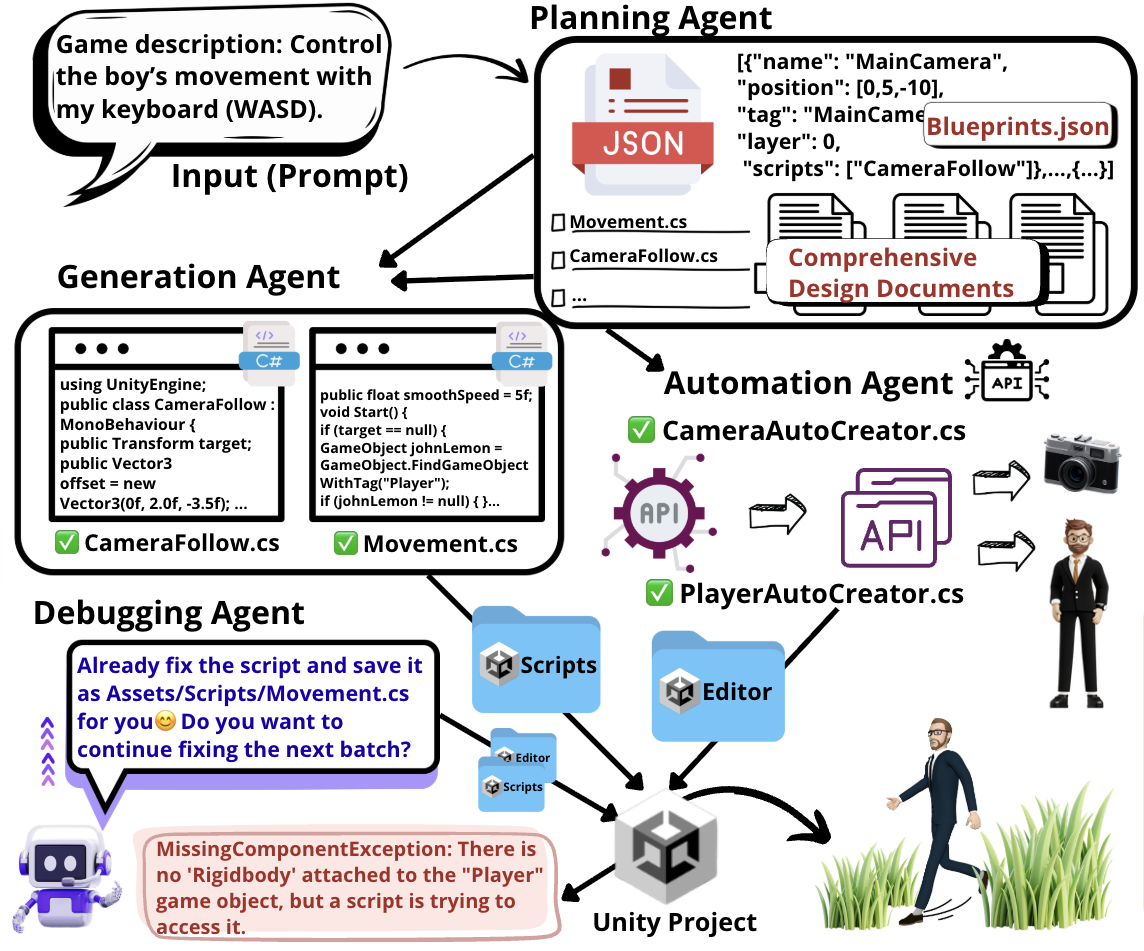}
    \vspace{-1.5em}
    \caption{Overview of \methodname, the four agents are powered by MLLMs}
    \vspace{-1.5em}
    \label{fig:unigen}
\end{figure}

\textbf{Overview.} Figure~\ref{fig:unigen} presents \methodname, a coordinated multi-agent MLLM framework that transforms natural language requirements into fully functional Unity projects.
Unlike existing approaches that generate only partial components or require extensive manual integration, \methodname automates the entire development lifecycle from high-level planning to engine-specific debugging.

The framework specifically addresses three primary challenges in current AI-assisted game development: 
(1) restriction to 2D content generation or isolated code snippets; 
(2) dependency on manual integration of generated components into game engines; and 
(3) inadequate handling of interactive logic and state management. 
Through sequential collaboration of four specialized agents, i.e., Planning, Generation, Automation, and Debugging, \methodname achieves end-to-end automation of 3D game development.

\textbf{Planning Agent\footnote{All MLLM prompt templates
are available 
at \url{https://github.com/yxwan123/UniGen}.}}. 
The Planning Agent 
interprets natural language requirements and decomposes them into structured game design specifications.
Its core component, 
the blueprint generator, produces a JSON blueprint that defines game mechanics, entities, attributes, and interactions.
This blueprint 
serves as the canonical reference for all subsequent generation steps, ensuring alignment with the original intent of the user.
Planning Agent enhances consistency 
through two key mechanisms.
First, it 
predefines uniform function names and component configurations to ensure coherent script generation. 
Second, the description generator creates detailed logic descriptions that effectively produce comprehensive design documents before code generation begins.

\textbf{Generation Agent.}
The agent translates structured blueprints into executable Unity C\# scripts through its enhanced C\# script generator.
This generator parses game functionality specifications and applies standardized implementation patterns for common mechanics, 
including \texttt{UIManager} integration, camera follow systems, player movement controllers, and automatic NPC path generation. 
The agent enforces strict output requirements that ensure adherence to Unity best practices. By incorporating proven design patterns and explicit code quality constraints, it produces scripts that require minimal manual adjustment while maintaining functional consistency across diverse game prototypes. 
This systematic approach to code generation bridges the gap between high-level specifications and production-ready implementation.

\textbf{Automation Agent.}
This agent addresses the critical engine implementation gap by creating a playable scene within the Unity Editor using generated scripts and blueprints as input. Its core component, the Editor Script Generator, has been upgraded with detailed Unity Editor API examples and structured prompts to produce robust custom editor scripts. 
A key innovation is the integration of the \texttt{ReflectionHelper} utility class, which uses reflection mechanisms to configure script fields during the automated scene assembly process safely. This approach prevents common runtime errors caused by null references or type mismatches, ensuring type safety and functional integrity. The generated editor scripts programmatically instantiate GameObjects, attach components, and correctly assign the generated C\# scripts, thus transforming the blueprint specifications into a fully initialized and playable game scene.

\textbf{Debugging Agent.}
The Debugging Agent provides iterative error correction through a conversational interface.
Developers 
describe specific errors or issues in natural language, and then the issues will be analyzed by the Debugging Agent.
The agent's key capability is simultaneous multi-script modification,
ensuring consistency across interrelated components. 
By automatically updating all affected files with coordinated fixes, the agent eliminates manual code editing while maintaining functional integrity throughout the debugging process.

\textbf{Implementation.}  
\methodname implements a structured Unity project workflow where all generated assets are systematically organized within the standard \textit{Assets} folder. The system segregates output into \textit{Editor} scripts for scene construction and \textit{Runtime} scripts for gameplay logic, enabling direct project execution without manual configuration. 
Our implementation of \methodname adheres to three core design principles: 
(1) Primitive-first 3D design, utilizing basic Unity shapes like Cubes and Spheres to ensure evaluability and controllability; 
(2) Functional grounding, establishing deterministic mappings between natural language intents (like “press W to move forward”) and Unity implementations (like script logic and component settings); and 
(3) Scene coherence, maintaining consistency through consistent naming, dependency resolution, and hierarchy management. 
For the evaluations in Section~\ref{sec:experiment}, we chose GPT-4.1\footnote{The MLLM backbone can be configured by \methodname's users flexibly.}~\cite{openai2025gpt41} as the MLLM backbone for its state-of-the-art code generation capabilities and robust reasoning in complex multi-step tasks.

\section{Experimental Evaluations}
\label{sec:experiment}
To rigorously assess 
\methodname's effectiveness and generalizability, we conducted 
evaluations focusing on two primary research questions: 
(1) the functional completeness across diverse genres and complexity levels, and (2) the quantitative efficiency improvements compared to traditional manual development approaches.

We evaluated \methodname on three distinct game prototypes to test its effectiveness and robustness. These games were chosen to represent a diverse set of genres and complexity levels, ensuring comprehensive validation across different scenarios:
    (1) \textbf{Test game \#1: Obstacle Run~\cite{unity_runner_template}.} A 3D platformer requiring a player-controlled ball to navigate a course with obstacles to reach an endpoint, testing basic movement, collision detection, and win/lose conditions.
    (2) \textbf{Test game \#2: Coin Collection~\cite{fulton_coin_collector}.} A game focusing on object interaction and state management, where a ball collects coins on a plane, testing triggers, scoring, and UI updates.
    (3) \textbf{Test game \#3: John Lemon's Haunted Jaunt~\cite{unity_johnlemon_2019}}. A more complex room escape game utilizing third-party assets, testing the system's ability to integrate external resources and manage more intricate object interactions and logic.

For each game, we constructed an evaluation matrix specifying important gameplay interactions (e.g., "Ball - touch barrier - game over"). Each interaction was evaluated using binary scoring (1 for correct implementation, 0 for failure) on the final generated executable. The overall \textit{functional completeness} was calculated as the percentage of successfully implemented interactions relative to the total specified interactions. 
Table~\ref{tab:func-evaluation} presents the functional completeness results across all three games.
\methodname achieved high scores of 100.0\% (15/15), 93.8\% (15/16), and 89.5\% (17/19) for the three games respectively.
These results demonstrate \methodname's robust zero-coding development performance across varying genres and complexity levels.
\begin{table}[t!]
\centering
\caption{Functional completeness evaluation of \methodname on benchmark Games. ``I'' stands for Interactions.}
\vspace{-1em}
\label{tab:func-evaluation}
\resizebox{0.9\columnwidth}{!}{%
\begin{tabular}{lccc}
\toprule
\textbf{Game} & \textbf{\# of Defined I} & \textbf{\# of Successful I} & \textbf{Functional Completeness} \\
\midrule
\# 1 & 15 & 15 & 100.0\% \\
\# 2 & 16 & 15 & 93.8\% \\
\# 3 & 19 & 17 & 89.5\% \\
\bottomrule
\end{tabular}
}
\vspace{-1em}
\end{table}

To assess development efficiency, we conducted a controlled comparison between \methodname 
traditional manual development.
For the manual baseline, we recruited undergraduate computer science student with more than three years of development experience to implement the same game following official Unity tutorials~\cite{unity_johnlemon_2019}.
During the development, we recorded their development time and the number of manual operations performed (e.g., script attachments, component configurations, parameter adjustments).

Table~\ref{tab:dev-comparison} shows the comparative results.
\methodname presents a statistically significant 91.4\% reduction in development time,
from approximately 140 minutes to under 12 minutes. 
Additionally, \methodname reduced manual operations by 93.3\%,
virtually eliminating repetitive tasks such as script attachment and component configuration that consumed substantial developer effort in the traditional approach.
The 91.4\% reduction in development time for basic prototypes represents a paradigm shift in rapid prototyping capabilities.
\methodname's parallelized architecture suggests even greater benefits for larger-scale projects.
This aligns with the industry trend, as generative AI is increasingly reshaping the game development process~\cite{UnityGamingReport2024}. 
\methodname automates technical implementation, allowing developers to focus more on creative design, accelerate prototyping, and lower the skill barrier for 3D game development. Its success in managing core gameplay loops and asset integration, shown by high functional completeness scores, underscores its potential for indie developers, educators, and rapid prototyping.

\begin{table}[t!]
\centering
\caption{Comparison of development efficiency between manual and \methodname-assisted development.}
\vspace{-1em}
\label{tab:dev-comparison}
\resizebox{\columnwidth}{!}{%
\begin{tabular}{lcccc}
\toprule
\textbf{Metric} & \textbf{Manual} & \textbf{\methodname} & \textbf{Improvement} & \textbf{P-Value} \\
\midrule
Development Time (mins) & $140$ & \textbf{$\leq12$} & 91.4\% $\downarrow$ & $<0.001$ \\
Manual Operations          & $75$ & \textbf{$\leq5$}             & 93.3\% $\downarrow$ & N/A \\
\bottomrule
\end{tabular}
}
\vspace{-1em}
\end{table}

\section{Limitations and Future Work}
While \methodname demonstrates robust capabilities in automating 3D game development, including instantiating pre-existing assets, characters' animators, components, and scripts, and implementing fundamental interactive logic, it currently exhibits limitations when handling more complex game mechanics. Specifically, \methodname shows constraints in autonomously generating sophisticated NPC behavior trees and managing the synchronization logic required for multiplayer gameplay. These limitations stem from the inherent challenges in modeling dynamic decision-making processes and real-time networked interactions through purely automated pipelines. Our future work will prioritize enhancing the system's capacity in these advanced domains. 

\section{Conclusion}
This work demonstrates that \methodname, our multi-agent framework integrating modern MLLMs,
can successfully automate substantial portions of 3D game development 
through a coordinated pipeline of Planning, Generation, Automation, and Debugging Agents. The framework effectively bridges natural language prompts with executable Unity projects via robust engine integration.
Experimental results across diverse game prototypes show \methodname reduces development time by 91.4\% on average while achieving functional completeness scores above 89.5\%, confirming its potential to democratize game creation by lowering technical barriers. 
Future work will expand support for complex game mechanics and advance end-to-end automation in interactive content creation.
\bibliographystyle{ACM-Reference-Format}
\bibliography{sample-base}

\end{document}